# Object-based Metamorphic Testing through Image Structuring

Adrian Wildandyawan
School of Informatics and Engineering
The University of Electro-Communications
Tokyo, Japan
Adrian.Wildandyawan@gmail.com

Yasuharu NISHI
School of Informatics and Engineering
The University of Electro-Communications
Tokyo, Japan
Yasuharu.Nishi@uec.ac.jp

*Abstract*—Testing software is often costly due to the need of mass-producing test cases and providing a test oracle for it. This is often referred to as the oracle problem. One method that has been proposed in order to alleviate the oracle problem is metamorphic testing. Metamorphic testing produces new test cases by altering an existing test case, and uses the metamorphic relation between the inputs and the outputs of the System Under Test (SUT) to predict the expected outputs of the produced test cases. Metamorphic testing has often been used for image processing software, where changes are applied to the image's attributes to create new test cases with annotations that are the same as the original image. We refer to this existing method as the image-based metamorphic testing. In this research, we propose an object-based metamorphic testing and a composite metamorphic testing which combines different metamorphic testing approaches to relatively increase test coverage.

## I. Motivation

Testing is an integral part of software engineering as its aim is to detect various bugs in a software. However, it requires a considerable amount of resources to create test cases. This includes preparing a test oracle to judge whether the system under test (SUT) produced the correct output for a given test case. In most cases, humans are required to manually assign the expected result, known as the test oracle, for test cases. The cost of this process is called the human oracle cost [1, 2]. Because of this, it can be difficult to provide a test oracle for verifying the result of a software test. This is referred to as the test oracle problem [3].

Metamorphic testing [4] is a popular testing method used to alleviate the test oracle problem. It produces new test cases from existing ones. Many metamorphic testing assign the same test oracle as the original test case to the new ones. Others assign oracles that are easily deducible from the original test case's. The relation used to deduce an oracle from another oracle is called the metamorphic relation.

Metamorphic testing can also be used for image processing software, where test cases are in the form of images, and the expected outputs are stored in the form of annotations [5, 6, 7]. New images can be generated from the existing images by altering the attributes thereof, such as blurring [6], changing the rotation [5, 6, 7], and so on. The annotations of the new images are the same as the original image. We refer to this existing approach as the image-based metamorphic testing.

In this research, we consider a different point of view for conducting metamorphic testing. As an image is fundamentally a graphical representation of certain objects, we decide to specifically focus on the objects within an image. Furthermore, because the identity of the image is decided by the structure of the objects, we consider applying alterations to the object's structure. We refer to this approach as the object-based metamorphic testing.

Moreover, we propose a composite metamorphic testing which combines different metamorphic testing approaches— such as image-based and object-based metamorphic testing— to improve test coverage and the type of detected misrecognition. Lastly, we conduct a case study using classifiers trained with the MNIST dataset.

## II. Related Works

### A. Deep Neural Network

Deep Neural Network (DNN) is a branch of deep learning that uses artificial neural network, modeled loosely after how the neuron works in living organisms. A DNN consists of three main layers—an input layer, multiple hidden layers, and an output layer that consist of connected nodes. Each node connection has its own weight which is used when a DNN classifier predicts an output for an input data.

A DNN classifier is trained to classify inputs by extracting features from the training data, and develop a set of weights for each node connection. Each layer of nodes recognizes features at different level of abstractions, where more complex features are recognized in layers closer to the output layer [8].

### B. AI-OCR

An optical character recognition (OCR) software is used to identify scanned physical characters and convert them into electronic characters. Traditional OCR software recognizes characters by using pattern-matching algorithms where the input image is compared to stored patterns pixel by pixel to predict the character. Because of this, the traditional OCR software lacks accuracy in recognizing handwritings. In comparison, AI-OCR software uses neural-network-based models that have been trained to recognize both printed and handwritten characters.

## C. MNIST Dataset

MNIST [9] is a dataset of handwritten digits, containing 60,000 images for training, and 10,000 images for validating. The handwritten digits are size-normalized to fit in a 20x20 pixel box while still maintaining the original aspect ratio. The center of mass of the size-normalized digits are then calculated and used to center each digit in a 28x28 pixel image.

There has been various classifiers trained using the MNIST training dataset with different classification architectures, such as the 2-hidden-layer neural network, Convolutional net LeNet-5 [10], and VGG16 [11]. 2-hidden layer neural network, as the name suggests, is an architecture comprised of, excluding the last fully connected layer, only 2 hidden layers. This is comparatively small in comparison to LeNet-5 and VGG16 architecture which uses each 6 hidden layers and 20 hidden layers.

## D. Metamorphic Testing

Metamorphic testing [4] is a testing method where new test cases are produced by altering an existing test case of which the output is correctly predicted by the SUT. The oracle or annotation of the produced test case is predicted based on its metamorphic relation to the original test case's annotation.

In the case of image processing software, a new image is produced by applying alteration to the original image in a way that keeps the annotation of the produced image the same with that of the original image.

The existing approach of metamorphic testing is an image-based metamorphic testing [5, 6, 7]. This approach treats the input data as images with attributes, and produce new test cases by altering these attributes, e.g., by changing the degree of rotation [5, 6, 7], blurring [6], shearing [6], and applying noise [7].

## III. PROPOSAL

In order to detect bugs that cannot be detected by the existing metamorphic testing, we propose an object-based approach of metamorphic testing. Furthermore, we propose a composite metamorphic testing that combines different metamorphic testing approaches.

### A. Object-based Metamorphic Testing

Object-based metamorphic testing handles the input data as objects with a structure, and new images are produced by altering the structure of the objects.

In the case of AI-OCR software, the characters are treated as objects of which the structure is composed of line segments. New images are produced by changing, e.g., the length, the degree of rotation, and the position of the line segments relative to the whole structure.

In this paper, we alter only the length of the line segments to produce new images from an existing image. We perform this through the steps shown on Figure 1.

### B. Composite Metamorphic Testing

Composite metamorphic testing combines different approaches of metamorphic testing in order to increase the overall test coverage.

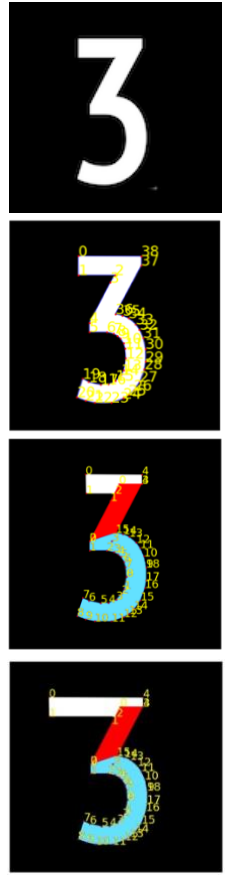

Step 1. The original image is preprocessed to center the digit within the image.

Step 2. The outline of the digit is transformed into an array of coordinate points.

Step 3. The coordinate points are grouped based on each individual line segment.

Step 4. The length of the line segment is altered by changing the coordinate position of the constituting coordinate points.

Fig. 1. Process of changing the length of a line segment by altering the coordinate position of its coordinate points.

In this research, test cases from image-based metamorphic testing and object-based metamorphic testing are combined to carry out a composite metamorphic testing. Image-based metamorphic testing is useful to detect misrecognitions caused by image capturing devices, e.g., image scanner and camera. Object-based metamorphic testing is useful to detect misrecognitions of handwritten digits. Because the two approaches of metamorphic testing are designed to detect different kinds of misrecognitions, combining the two approaches will boost the test coverage in comparison to each of the individual approach.

## IV. EXPERIMENTAL EVALUATION

In this paper, we experimentally conduct the image-based metamorphic testing, the object-based metamorphic testing, and the composite metamorphic testing of the two approaches. The experiment is carried out on Google Colaboratory with the following specifications:

- NVIDIA GPU : Tesla T4
- GPU RAM : 12.0 GB
- CPU Chip : Intel® Xeon® CPU @ 2.00GHz
- Total CPU RAM : 13.3 GB

### A. Training Classifiers

In this experiment, we train 3 classifiers with MNIST training dataset. The architecture of these classifiers are based on 2-hidden layer Neural Network (NN) classifier [10], LeNet-5 [10], and VGG16 [11] as shown in Table I.

TABLE I. ARCHITECTURES OF THE THREE CLASSIFIERS [10,11]

| 2-HL NN based | LeNet-5 based | VGG16 based |
|---|---|---|
| Input data (Flattened 28 x 28 greyscale image) | Input data (28 x 28 greyscale image) | Input data (28 x 28 greyscale image) |
| Dense-512 | Conv2D-6 | Conv2D-8 |
| | | Conv2D-8 |
| | AvgPool | MaxPool |
| | | Conv2D-16 |
| | Conv2D-16 | Conv2D-16 |
| Dense-512 | | MaxPool |
| | AvgPool | Conv2D-32 |
| | | Conv2D-32 |
| | Dense-120 | Conv2D-32 |
| | | MaxPool |
| Dense-10 | Dense-84 | Dense-1024 |
| | | Dense-1024 |
| | Dense-10 | Dense-10 |
| Softmax | Softmax | Softmax |

### B. Image-based Metamorphic Testing

For the image-based metamorphic testing, we choose the font "PT Sans Narrow" to create 10 images of digit 0 to 9 to use as the original image. This font is used because all the classifiers managed to correctly predict each image produced with this font, which is a requirement for metamorphic testing.

We decide to combine 3 levels of Gaussian blur in the x-axis direction, 3 levels of Gaussian blur in the y-axis direction, 3 levels of Gaussian noise, 3 levels of shearing, 3 levels of rotation, and 3 levels of erosion. This results in a total of 7290 altered images.

### C. Object-based Metamorphic Testing

For the object-based metamorphic testing, we also use the 10 images created by using the font "PT Sans Narrow." We extract different numbers of line segments for each digit, and produce new images by changing the length of each line segment on both ends as shown in Table II. This results in a total of 3906 new images.

### D. Composite Metamorphic Testing

For the composite metamorphic testing, we combine the images produced with image-based and object-based metamorphic testing and carry out a test with them.

TABLE II. NUMBER OF LINE SEGMENTS THAT EACH DIGIT IS DIVIDED INTO AND NUMBER OF VARIATIONS OF ELONGATION FOR EACH LINE SEGMENT.

| Digit | Number of Line Segments | Number of Variations of Elongation |
|---|---|---|
| 0 | 2 | 3 |
| 1 | 3 | 3 |
| 2 | 2 | 3 |
| 3 | 3 | 3 |
| 4 | 3 | 3 |
| 5 | 3 | 3 |
| 6 | 1 | 3 |
| 7 | 3 | 3 |
| 8 | 2 | 3 |
| 9 | 1 | 3 |

TABLE III. THE RESULT OF IMAGE-BASED METAMORPHIC TESTING.

| Classifier | Number of produced images | Number of detected misrecognitions | Rate of misrecognized image | Time taken (s) to produce images |
|---|---|---|---|---|
| 2-HL NN | 7290 | 1954 | 26.80% | 2521.80 s |
| LeNet-5 | | 1556 | 21.34% | |
| VGG16 | | 1582 | 21.70% | |
| Average | | 1697.3 | 23.28% | |

TABLE IV. THE RESULT OF OBJECT-BASED METAMORPHIC TESTING.

| Classifier | Number of produced images | Number of detected misrecognitions | Rate of misrecognized image | Time taken (s) to produce images |
|---|---|---|---|---|
| 2-HL NN | 3906 | 1515 | 38.79% | 570.66 s |
| LeNet-5 | | 1522 | 38.97% | |
| VGG16 | | 1514 | 38.76% | |
| Average | | 1517.0 | 38.84% | |

TABLE V. THE RESULT OF COMPOSITE METAMORPHIC TESTING.

| Classifier | Number of produced images | Number of detected misrecognitions | Rate of misrecognized image | Percentage of increase in comparison to image-based metamorphic testing | Time taken (s) to produce images |
|---|---|---|---|---|---|
| 2-HL NN | 11196 | 3469 | 30.98% | +77.53% | 3092.46 s |
| LeNet-5 | | 3078 | 27.49% | +97.81% | |
| VGG16 | | 3096 | 27.65% | +95.70% | |
| Average | | 3214.3 | 28.71% | +89.39% | |

## V. EXPERIMENTAL RESULTS

### A. Image-based Metamorphic Testing

The results of the image-based metamorphic testing on the three classifiers are summarized in Table III. The average number of misrecognitions detected from each classifier was 1697.3 images, which is 23.28% of the total number of produced images. Furthermore, the average time it takes to produce one image was 1.49 seconds. Several examples of the images that are misrecognized are shown in Figure 2.

### B. Object-based Metamorphic Testing

The results of the image-based metamorphic testing on the three classifiers are summarized in Table IV. The average number of misrecognitions detected from each classifier was 1517 images, which is 38.84% of the total number of produced images. Furthermore, the average time it takes to produce one image was 0.38 seconds. Several examples of the images that are misrecognized are shown in Figure 3.

### C. Composite Metamorphic Testing

The results of the image-based metamorphic testing on the three classifiers are summarized in Table V. The average number of misrecognitions detected from each classifier was 3214.3 images, which is 28.71% of the total number of produced images. This equates to an average of 89.39% increase of misrecognitions in comparison to only using image-based metamorphic testing.

Because the image-based and object-based metamorphic testing produced images that are nonidentical to ones produced by the other approach, we could consider that the images produced through the composite metamorphic testing are the combination of images produced through image-based and object-based metamorphic testing.

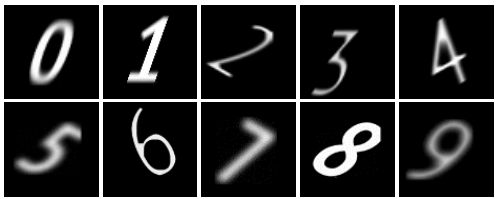

Fig. 2. Images produced through image-based metamorphic testing.

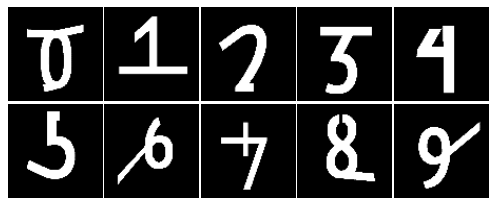

Fig. 3. Images produced through object-based metamorphic testing.

## VI. DISCUSSIONS

Through composite metamorphic testing, we were able to combine the result of the image-based and object-based metamorphic testing. Because the image-based and object-based metamorphic testing produced different kinds of images, even though there is no method to objectively measure the test coverage of metamorphic testing, it can be said that the composite metamorphic testing was able to achieve higher test coverage relative to that of the other two approaches separately.

In this research, we apply object-based metamorphic testing for test cases containing a single object. However, this approach can also be applied to ones with multiple objects. In this case, alterations on each individual object can be combined to produce more test cases compared to only altering one individual object.

A threat to conclusion validity of this research would be whether this research could still be applied when the number of images required to be produced is scaled up. In order to prove the conclusion validity of this research, we shall prove the scalability of this method.

More number of images may be required to be produced for testing. Assuming that using each of metamorphic testing approach, 50,000 images are to be produced for a total of 100,000 new images, and that the time required to produce images increase linearly, the time required to execute this would be 6.83 hours.

In comparison to this, we conducted a separate experiment where 4 subjects were required to manually produce images for testing, each within a 40-minute timeframe. As a result, on average each subject were able to produce 456 images, 20.04% of which were misrecognized by each model. Using the same assumption above, the average time required for a subject to manually produce 100,000 images would be 146.20 hours.

The proposed method of producing new images does require more time as the number of images to be produced increases tenfold, but the time required to do so is still 95.32% lower than the time it would take a person to produce new images manually. Therefore, we were able to prove the scalability of this method, and thus proving the conclusion validity of this research.

Furthermore, a threat to external validity of this research would be whether the images were able to be produced only due to the GPU used in the testing environment. In order to prove the external validity of this research, we conducted the same research on the same Google Colaboratory environment without using GPU.

Firstly, we conducted the same experiment and recorded the time required to produce the same number of images under an environment with no GPU. Next, assuming that the time required to produce new images increase linearly, we plot a graph showing the required time when the number of images are scaled up. The result is shown in Figure 4.

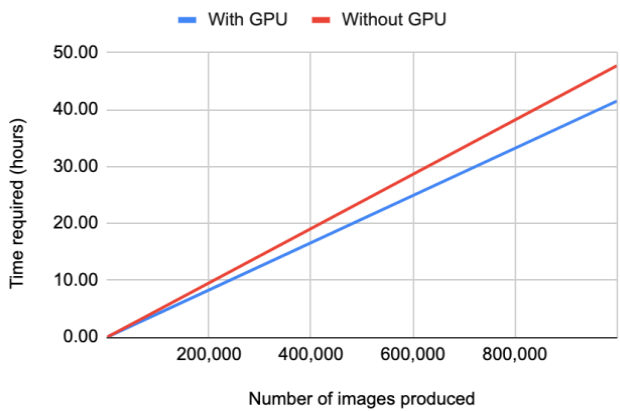

Fig. 4. Several of the images produced through object-based metamorphic testing.

As we can see, it requires more time to produce the same number of images without using GPU in comparison to producing them using GPU. However, the time it takes to produce images without GPU is only 15.00% longer compared to the time it takes to produce the same images with GPU. Additionally, when we scale up the number of images to be produced to 1,000,000 images, the time it takes when no GPU is used is only 47.69 hours in comparison to 41.46 hours when using GPU—both of which are significantly lower than the time it would take subjects to manually produce 1/10 of those images.

For this reason, we have proved that even though the system performs relatively slower when no GPU is used, it still outperforms manual method, therefore proving the external validity of this research.

## VII. Conclusion and Future Work

In this paper, we proposed object-based metamorphic testing that treats input data as objects and manipulate the data's structure to create new test cases, e.g. the line segments of digits by changing the length thereof. This approach, combined with the current image-based approach, made composite metamorphic testing possible. Through the experiments conducted, composite metamorphic testing were able to detect more misrecognized images compared to the individual metamorphic testing approaches, therefore relatively increasing the test coverage of metamorphic testing.

There are several possible future works that can be done in continuation of this research. First, even though we only employ changing the length of line segments of images in this research, other methods can also be used, e.g. changing the rotation, curvature, and position of the line segments. Secondly, as the approach of object-based metamorphic testing and composite metamorphic testing is not exclusive to image recognition for numbers, this research can also be applied to test various other DNN-based software.

## References

[1] P. McMinn, M. Stevenson and M. Harman, "Reducing Qualitative Human Oracle Costs associated with Automatically Generated Test Data," pp. 1-4, 2010.

[2] E. T. Barr, M. Harman, P. McMinn, M. Shahbaz and S. Yoo, "The Oracle Problem in Software Testing: A Survey," *IEEE Transactions on Software Engineering,* vol. 41, no. 5, pp. 507-525, 2015.

[3] W. E. Howden, "Theoretical and Empirical Studies of Program Testing," *IEEE Transactions on Software Engineering,* vol. 4, no. 4, pp. 293-298, 1978.

[4] T. Y. Chen, S. C. Cheung and S. W. Yiu, "Metamorphic testing: a new approach for generating next test cases," 1998.

[5] A. Dwarakanath, M. Ahuja, Sikand, Samarth, R. M. Rao, R. P. J. C. Bose, N. Dubash and S. Podder, "Identifying Implementation Bugs in Machine Learning Based Image Classifiers using Metamorphic Testing," *ISSTA 2018: Proceedings of the 27th ACM SIGSOFT International Symposium on Software Testing and Analysis,* pp. 118-128, 2018.

[6] H. Spieker and A. Gotlieb, "Adaptive Metamorphic Testing with Contextual Bandits," 2 January 2020. [Online]. Available: https://arxiv.org/pdf/1910.00262.pdf. [Accessed 8 February 2020].

[7] R. R. Mekala, M. E. Gudjon, A. Porter, M. Lindvall and M. Diep, "Metamorphic Detection of Adversarial Examples in Deep Learning Models With Affine Transformations," in *MET '19: Proceedings of the 4th International Workshop on Metamorphic Testing*, 2019.

[8] C. Molnar, Interpretable machine learning. A Guide for Making Black Box Models Explainable, 2019.

[9] Y. LeCun, C. Cortes and C. J. Burges, "The MNIST Databse of Handwritten Digits," [Online]. Available: http://yann.lecun.com/exdb/mnist/. [Accessed 7 February 2020].

[10] Y. LeCun, L. Bottou, Y. Bengio and P. Haffner, "Gradient-Based Learning Applied to Document Recognition," *Proceedings of the IEEE,* vol. 86, no. 11, pp. 2278-2324, 1998.

[11] K. Simonyan and A. Zisserman, "Very Deep Convolutional Networks for Large-Scale Image Recognition," in *arXiv 1409.1556.*, 2014.